\documentclass{article} 
\usepackage{iclr2024_conference,times}

\usepackage{amsmath,amsfonts,bm}

\def\eqref#1{equation~\ref{#1}}

\def\1{\bm{1}}

\DeclareMathAlphabet{\mathsfit}{\encodingdefault}{\sfdefault}{m}{sl}
\SetMathAlphabet{\mathsfit}{bold}{\encodingdefault}{\sfdefault}{bx}{n}

\usepackage{graphicx} 
\usepackage{hyperref}
\usepackage{url}
\usepackage{array}
\usepackage{booktabs}
\usepackage{multirow}

\title{Seeing Through the Clouds: Cloud Gap Imputation with Prithvi Foundation Model}

\author{Denys Godwin, Hanxi (Steve) Li, Michael Cecil \& Hamed Alemohammad\\
Graduate School of Geography and Clark Center for Geospatial Analytics\\
Clark University\\
Worcester, MA 01610, USA \\
\texttt{dgodwin@clarku.edu} \\
}

\iclrfinalcopy 
\begin{document}

\maketitle

\begin{abstract}
Filling cloudy pixels in multispectral satellite imagery is essential for accurate data analysis and downstream applications, especially for tasks which require time series data. To address this issue, we compare the performance of a foundational Vision Transformer (ViT) model with a baseline Conditional Generative Adversarial Network (CGAN) model for missing value imputation in time series of multispectral satellite imagery. We randomly mask time series of satellite images using real-world cloud masks and train each model to reconstruct the missing pixels. The ViT model is fine-tuned from a pretrained model, while the CGAN is trained from scratch. Using quantitative evaluation metrics such as structural similarity index and mean absolute error as well as qualitative visual analysis, we assess imputation accuracy and contextual preservation. 
\end{abstract}

\section{Introduction}

The enormous scale of satellite image datasets along with the proliferation and sophistication of machine learning models has led to the development of Geospatial Foundation Models (GFM) which use self-supervised learning to encode patterns from terabytes of data for which high-quality ground truth labels do not exist (e.g. \cite{lacoste2023geobench, jakubik2023foundation, klemmer2023satclip, tseng2024lightweight, cha2023billion, satmae2022}). These patterns of encoded spatial, temporal, and band relationships can then be decoded with decoder heads that can be fine-tuned for individual use-cases. This not only has the potential to reduce the training time, cost, and resource use of machine learning inference, but also to improve model performance.

Self-supervision for satellite imagery may be achieved using masked autoencoders learners \citep{he2021masked}. Missing data imputation is therefore an inherent ability of FMs trained in this way. Given that Missing data due to clouds is a persistent issue in multispectral satellite imagery, self-supervised models that excel at cloud gap imputation and can transfer this ability to other geographic regions with little fine-tuning could assist in downstream tasks such as land use change detection or crop monitoring.

The utility of cloud gap imputation, whether done by artificial neural networks or by algorithmic interpolation methods, is dependent on its downstream application. Any reconstructed pixel value is a "best guess" based on learned representations of spatial and temporal context, and does not represent the true pixel value at a time and location. However, the augmentation of training datasets with synthetic data can improve the performance of computer vision models, as shown in the medical field by \citet{8756037}. Additionally, the accuracy of a generative model in pixel reconstruction is a useful metric in assessing its ability to encode deep representations of the input data.

In adapting the concept of FMs to satellite imagery, it is important to benchmark against current state-of-the-art models and assess the unique modalities and challenges of the remote sensing domain \citep{rolf2024mission}. Here, we measure the performance of Prithvi GFM against a Conditional Generative Adversarial Network (CGAN) for imputing cloud gaps in time series of Harmonized Landsat and Sentinel-2 (HLS) imagery using a range of sample sizes.

\section{Model Architectures}

Prithvi is a GFM based on the multi-temporal vision transformer (ViT) architecture (with 100M parameters) and has been trained on 1TB of multispectral satellite imagery from HLS dataset \citep{jakubik2023foundation}. The model is pretrained using a masked autoencoder learner on a dataset where each scene consists of 3 time steps of 6 bands each.  

\begin{figure}
\begin{center}
\fbox{\includegraphics[width=0.75\linewidth]{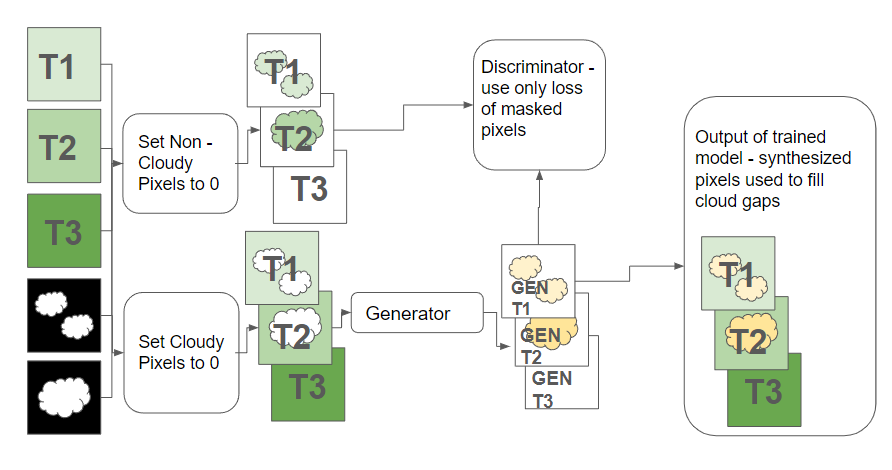}}
\end{center}
\caption{Training the CGAN to impute cloudy pixels was accomplished by masking out clouds from input data, using this as the condition on which to generate, then comparing the generated data against the unmasked ground truth.}
\label{cgan-diagram}
\end{figure}

The CGAN architecture contains two encoder-decoder convolutional neural networks which are trained against each other to generate realistic outputs based on input data. We use randomly masked time series scenes as the input condition (Figure \ref{cgan-diagram}). The CGAN model was adapted from \citet{Baier_2022} which is based on the landmark CGAN architecture Pix2Pix \citep{isola2017image}. We use a multi-scale patch-based discriminator to evaluate plausibility at two physical scales, for both small detail and wider spatial context. The use of the CGAN for inpainting gaps in input images is established using non-geospatial data \citep{demir2018patchbased}. This technique in the geospatial context benefits from our time-series data, as it allows values from each time step to inform guesses about missing values, thanks to the persistence, between different time scenes in a given year, of features on the ground. 

\section{Methodology}

In order to compare the performance of Prithvi to the baseline CGAN model, both models were trained using subsets of a cloud-free time-series dataset with randomly added real life cloud masks and tested on a reserved validation set which remained constant and identical across all experiments with no shuffling or random selection.

\subsection{Dataset and Experiment Setup}

Imagery from HLS dataset was collected across the Contiguous United States (CONUS) with diverse land cover classes (details in \cite{jakubik2023foundation}). Each chip covers a 224 x 224 pixel region, with a spatial resolution of 30 meters and six spectral bands from three temporal snapshots stacked together for 18 total channels. The three scenes are selected between Mar and Sep 2022 with time difference between scenes varying between 1 and 200 days. After filtering for missing values and cloudy pixels, a total of 7,852 cloud-free chips evenly distributed across the CONUS were generated. This set was randomly partitioned into 6,231 training chips and 1,621 validation chips. The same training and validation sets were used for all experiments. The training dataset was then reduced to subsets of 3,200, 1,600, 800, and 400 for successive experiments.

The dataset for these experiments was collected independently of the training dataset used to train Prithvi to prevent leakage. Prithvi was trained using 2017 HLS data, while these experiments were run using 2022 HLS data. While both datasets were randomly sampled from CONUS, sampling from different years ensures that image chips used in pretraining Prithvi are not part of these fine-tuning experiments.

Cloud masks were generated from the same region of CONUS using HLS cloud mask quality flag and exported as a binary layer of cloudy and non-cloudy pixels. This yielded 21,642 cloud masks, of which 1,600 were randomly selected and reserved for validation. The validation set was balanced such that there were an equal number of samples in each of 10 equal divisions between 1 and 100 percent coverage.

We setup two experiments: in experiment 1 (E1) only the middle scene in each chip was masked, and in experiment 2 (E2) all combination of all time steps were used for masking (e.g. all three scenes, or just the first two scenes, etc). All experiments were run for 200 epochs. 

\subsection{Prithvi Fine-Tuning}

The pretraining loop for Prithvi was modified to accept real-life cloud masks as inputs, replacing the random masking that had been used in training the model. As the ViT architecture only reconstructs missing patches, and incorporates no data from patches which are masked out, any patch which included a cloudy pixel was therefore removed entirely. The Mean Squared Error (MSE) of reconstructed patches as compared to their equivalent ground truth patches is used as the loss metric for training. In calculating metrics for comparison with the CGAN, only missing pixels are replaced with generated values, rather than replacing the entire patch that the missing pixels belong to.

\subsection{CGAN Training}

For CGAN training, first the generator weights are frozen and the discriminator is trained using hinge loss. Then, the discriminator weights are frozen and the generator weights are updated based on hinge loss compared to the discriminator outputs and to MSE calculated between true and generated values with hyperparameter $\alpha$ as the weight for MSE:

\begin{equation}
\text{Total Loss} = \text{hinge loss} + \alpha \cdot \text{MSE}
\end{equation}

MSE is only calculated for pixels that are masked out, and discriminator loss is only calculated for patches which contain at least one cloudy pixel. Figure \ref{cgan-diagram} shows how the CGAN was trained using cloud-masked time series as a condition and the ground truth as the target. For all experiments, discriminator learning rate was $1.0e-4$ and generator learning rate was $5.0e-4$, following \citet{heusel2018gans} and preliminary experiments in learning rate selection. $\alpha$ is set to $5$ for all experiments.

\section{Results and Discussion}

We use Structural Similarity Index Measure (SSIM) as a quantitative metric to compare outputs of Prithvi and CGAN against the ground truth, but SSIM is not used as a loss metric during training. SSIM was chosen to compliment pixel-wise Mean Absolute Error (MAE) as it is sensitive to changes in texture, spatial patterns, and contrast, which are important in downstream tasks such as semantic segmentation. Reconstructed time series, where generated values were imputed for cloud masks but all unmasked data was preserved, were compared against the original complete time series using SSIM. Unlike pixel-wise error, SSIM was not adjusted to account for cloud coverage, so values for this metric are highly dependent on cloud cover. 

In comparing results from E1 and E2 experiments, it is important to note that SSIM was calculated for only the center time step in E1, and across all time steps for E2. Therefore, SSIM between E1 and E2 experiments cannot be rigorously compared, and should serve only in comparing the results of Prithvi and CGAN model from the same round of experiments.

For E1 experiments, Table \ref{round1-table} and Figure \ref{round1-graphs} show that Prithvi outperforms CGAN in nearly all cases in terms of both MAE and SSIM. Only CGAN trained on the full training set marginally outperforms Prithvi with no fine-tuning. Table \ref{round1-vit-correlations} shows that Prithvi is also less affected by increases in time gap and cloudy percentage than the CGAN model, shown in Table \ref{round1-cgan-correlations}.

For E2 experiments, Prithvi outperforms the CGAN model as is shown in Table \ref{round2-table} and Figure \ref{round2-table}. Zero-shot inference by Prithvi achieves an MAE of 0.03 on all masked pixels in the validation dataset, which is reduced to below 0.025 in 10 epochs or under for all training data subsets. As mean pixel reflectance values for all bands of the validation dataset were $0.151$, MAE of $0.03$ would denote an average error of approximately 20 percent. 

Overall, the CGAN model consecutively shows improvement in both MAE and SSIM with larger subsets of data, suggesting that 6,231 scenes does not achieve saturation for the model. However, even with the full dataset, it is outperformed by Prithvi with no fine-tuning. Prithvi outperforms the CGAN despite being unable to exploit all available information due to discarding any input patch with any masked pixels during training and inference.

\begin{table}[h]
\centering
\caption{MAE and SSIM for best epoch of 200 for E2 experiments (applying masks in all time steps). Best epoch is the best performance out of 5 runs for all experiments across all epochs.}
\label{round2-table}
\begin{tabular}{*9c} 
    \toprule
    \multirow{3}{*}{Sample Size} &  \multicolumn{4}{c}{SSIM} &  \multicolumn{4}{c}{MAE} \\
     &  \multicolumn{2}{c}{Training} & \multicolumn{2}{c}{Validation} &  \multicolumn{2}{c}{Training} & \multicolumn{2}{c}{Validation}\\
     & Prithvi & CGAN & Prithvi & CGAN & Prithvi & CGAN & Prithvi & CGAN\\ 
    \midrule
    6231 & \textbf{0.949} & 0.919 & \textbf{0.960} & 0.937 & \textbf{0.024} & 0.033 & \textbf{0.022} & 0.032 \\ 
    3200 & \textbf{0.949} & 0.917 & \textbf{0.959} & 0.931 & \textbf{0.024} & 0.035 & \textbf{0.023} & 0.035 \\
    1600 & \textbf{0.950} & 0.916 & \textbf{0.958} & 0.926 & \textbf{0.024} & 0.035 & \textbf{0.023} & 0.037 \\
    800  & \textbf{0.950} & 0.905 & \textbf{0.957} & 0.916 & \textbf{0.024} & 0.038 & \textbf{0.024} & 0.042 \\
    400  & \textbf{0.947} & 0.894 & \textbf{0.956} & 0.912 & \textbf{0.025} & 0.045 & \textbf{0.024} & 0.045 \\
    0 (zero-shot) & \textbf{-} & - & \textbf{0.946} & - & \textbf{-} & - & \textbf{0.030} & - \\
    \bottomrule
\end{tabular}
\end{table}

In fine-tuning for E1, Prithvi experiences loss spikes before finding a new, higher performance local minimum, as is seen in Figure \ref{vit_training_graph_round1}. Similarly, for E2, Prithvi's best performance is achieved before 10 epochs on all fine-tuning experiments with a collapse in performance after that and a convergence to a new, suboptimal local minima (Figure \ref{vit_training_graph_round2}). This differs from fine-tuning E1 experiments because masking at multiple time steps is closer to the original pretraining method of Prithvi. 

\begin{figure}
\begin{center}
\fbox{\includegraphics[width=0.7\linewidth]{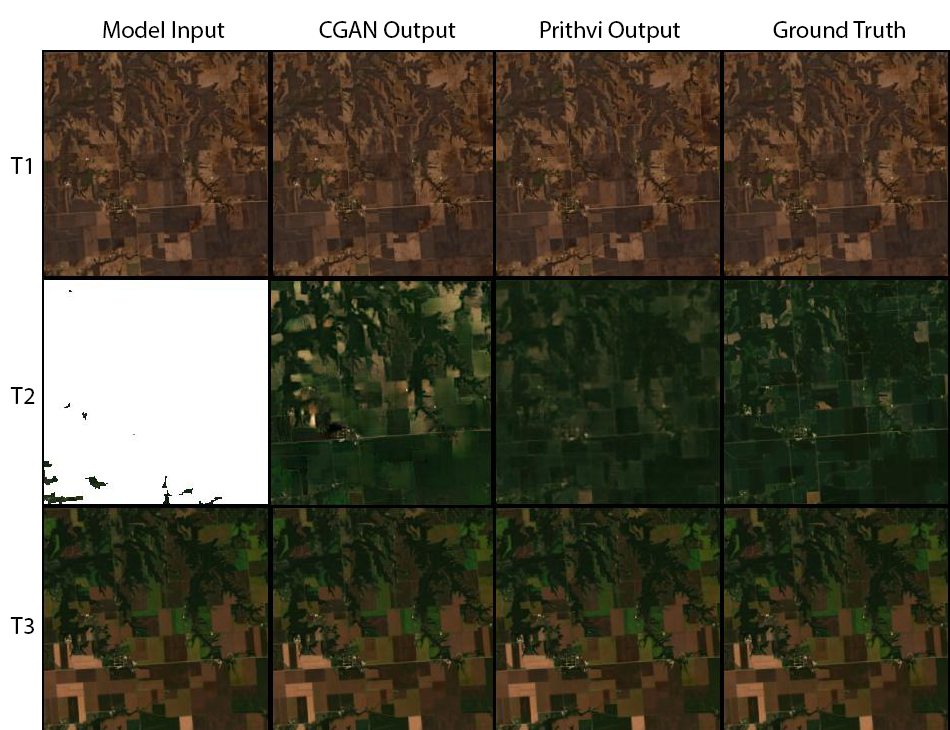}}
\end{center}
\caption{Reconstruction of a high-coverage image using CGAN and Prithvi, both trained using 6,231 images from E1 experiments (applying mask to the middle scene)}
\label{round1-fullcover}
\end{figure}

Visual inspection of true color reconstructed scenes show that both Prithvi and the CGAN are capable of filling cloudy pixels in challenging combinations of large time gaps and large chunks of missing data. The CGAN tends to preserve small details but  outputs unrealistic reflectance values and salt and pepper noise, especially where there is insufficient data. For example, in Figure \ref{round1-fullcover}, the CGAN outputs anomalously high reflectance values. Prithvi, on the other hand, does not tend to preserve details such as roads and buildings, but outputs reflectance values that are more constrained within the bounds of true observations. When there is insufficient data, the images are subject to a checkerboarding effect, as in Figure \ref{round2-fullcover}.

\section{Conclusion}

We find that Prithvi, even without fine-tuning, outperforms the CGAN on cloud gap imputation with all subsets of data. This shows the strength of GFMs on gap-filling tasks, given that this mimics the training loop of models such as Prithvi. The model's exposure to over 1TB of diverse images helps constrain its predicted values compared to the bespoke CGAN (Figures \ref{round2-cgan-correlation}-\ref{round2-vit-correlation-full}). The gap-filled imagery can be used to augment time series data for training other downstream applications which benefit from complete coverage and multi-temporal scenes (like crop type segmentation or crop yield estimation). Further research is needed to assess the extent to which cloud gap imputation as a data augmentation technique is effective in training better downstream models.

Given that the models perform better in E1 experiments (masked middle scene), we recommend this configuration for cloud gap imputation. Prithvi has no time encoding for scenes; therefore, non-cloudy scenes may be selected from across the range of the year.

One strength of time series satellite imagery for cloud gap imputation is the relative stationarity of features on the ground over time. This strength contributes to the ability of FMs to generate realistic pixel values in masked areas. Future research can incorporate non-imagery datasets such as Digital Elevation Models (DEMs) and land cover classification layers that can improve model performance.

\bibliography{iclr2024_conference}
\bibliographystyle{iclr2024_conference}

\appendix
\section{Appendix}

\begin{table}[h]
\centering
\caption{MAE and SSIM for best epoch of 200 for E1 experiments (applying mask in the middle scene).}
\label{round1-table}
\begin{tabular}{*9c} 
    \toprule
    \multirow{3}{*}{Sample Size} &  \multicolumn{4}{c}{SSIM} &  \multicolumn{4}{c}{MAE} \\
     &  \multicolumn{2}{c}{Training} & \multicolumn{2}{c}{Validation} &  \multicolumn{2}{c}{Training} & \multicolumn{2}{c}{Validation}\\
     & Prithvi & CGAN & Prithvi & CGAN & Prithvi & CGAN & Prithvi & CGAN\\ 
    \midrule
    6231 & \textbf{0.930} & 0.883 & \textbf{0.931} & 0.904 & \textbf{0.016} & 0.031 & \textbf{0.020} & 0.032 \\ 
    3200 & \textbf{0.933} & 0.872 & \textbf{0.931} & 0.896 & \textbf{0.016} & 0.033 & \textbf{0.021} & 0.035 \\
    1600 & \textbf{0.929} & 0.877 & \textbf{0.927} & 0.895 & \textbf{0.017} & 0.031 & \textbf{0.022} & 0.037 \\
    800  & \textbf{0.921} & 0.866 & \textbf{0.923} & 0.876 & \textbf{0.019} & 0.033 & \textbf{0.023} & 0.042 \\
    400  & \textbf{0.922} & 0.841 & \textbf{0.915} & 0.865 & \textbf{0.019} & 0.039 & \textbf{0.027} & 0.045 \\
    0 (zero-shot) & \textbf{-} & - & \textbf{0.887} & - & \textbf{-} & - & \textbf{0.030} & - \\
    \bottomrule
\end{tabular}
\end{table}

\counterwithin{figure}{section}

\begin{figure}[h]
\begin{center}
\fbox{
\includegraphics[width=6cm]{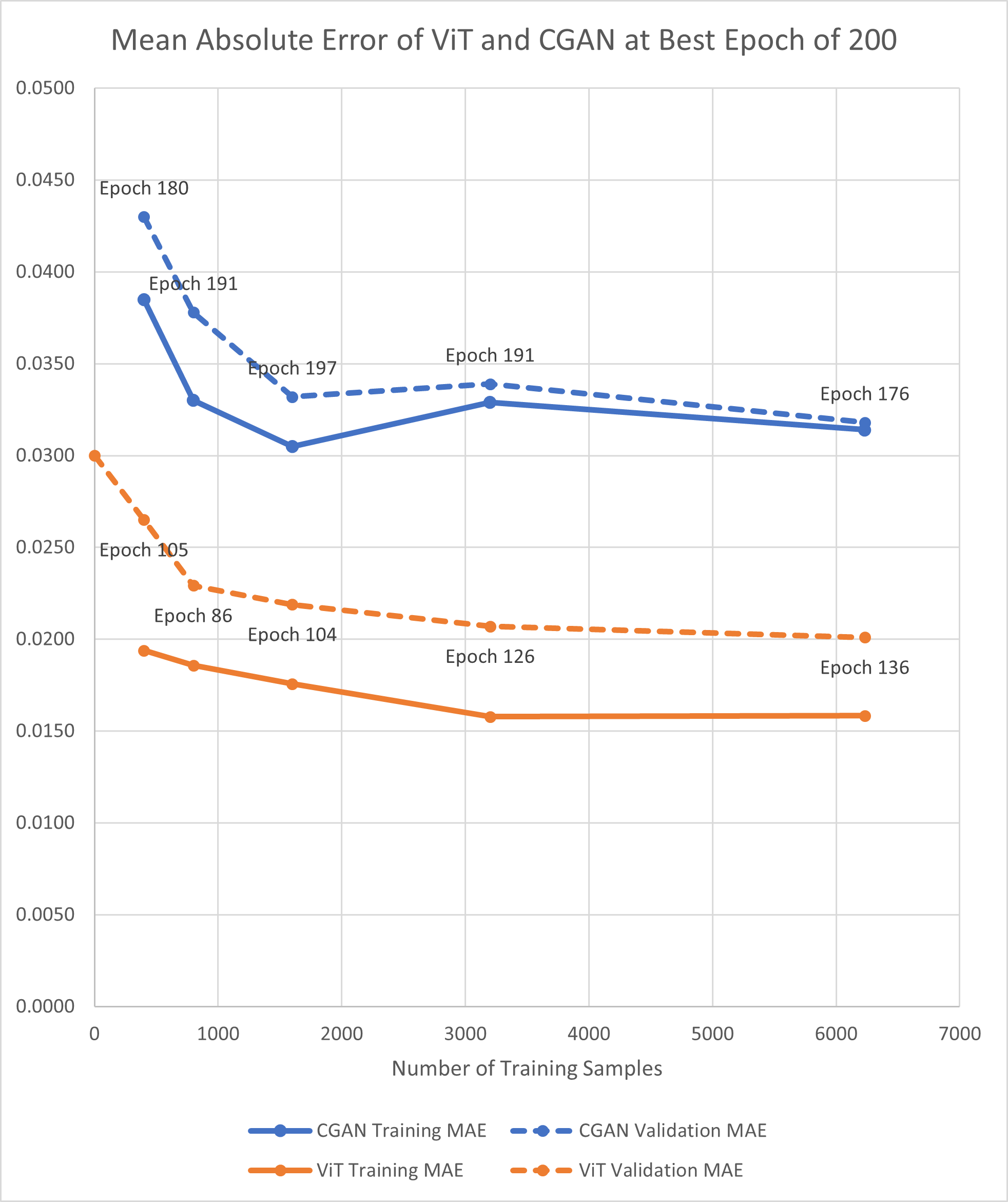}
\includegraphics[width=6cm]{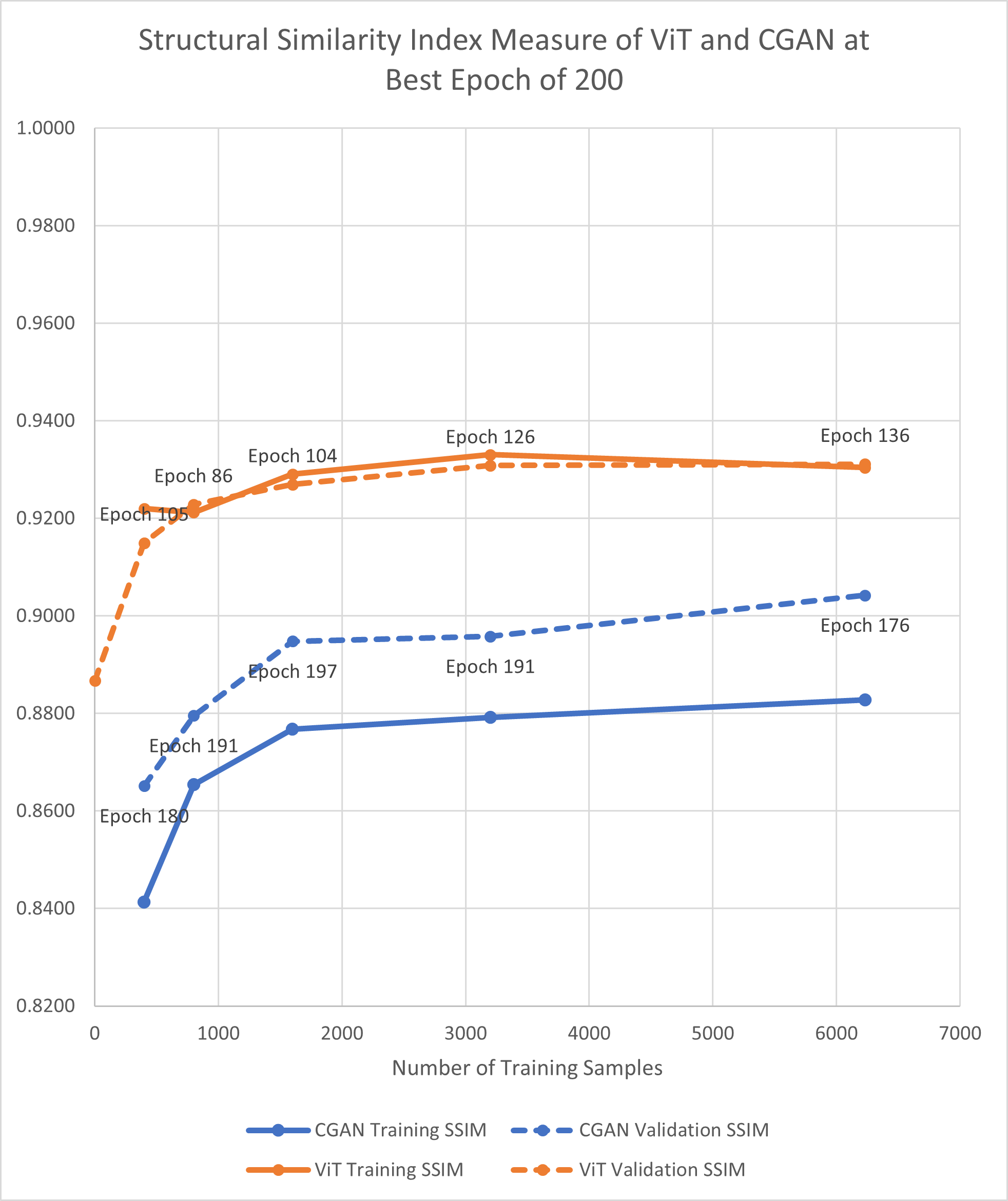}
}
\end{center}
\caption{MAE and SSIM for best epoch of 200 for E1 experiments (applying mask to the middle scene). Best epoch is the best performance out of 5 runs for all experiments across all epochs.}
\label{round1-graphs}
\end{figure}

\begin{figure}
\begin{center}
\fbox{
\includegraphics[width=6cm]{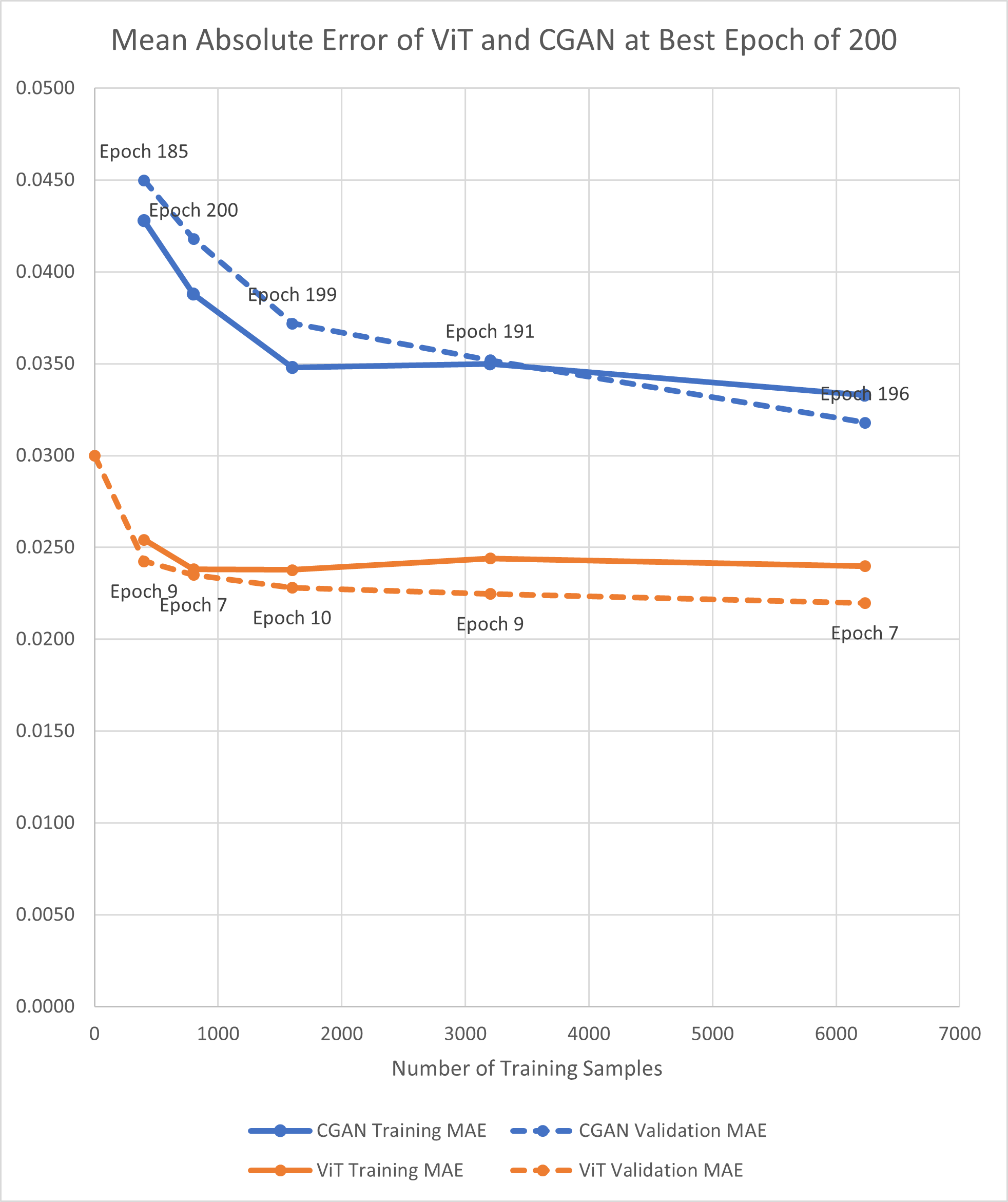}
\includegraphics[width=6cm]{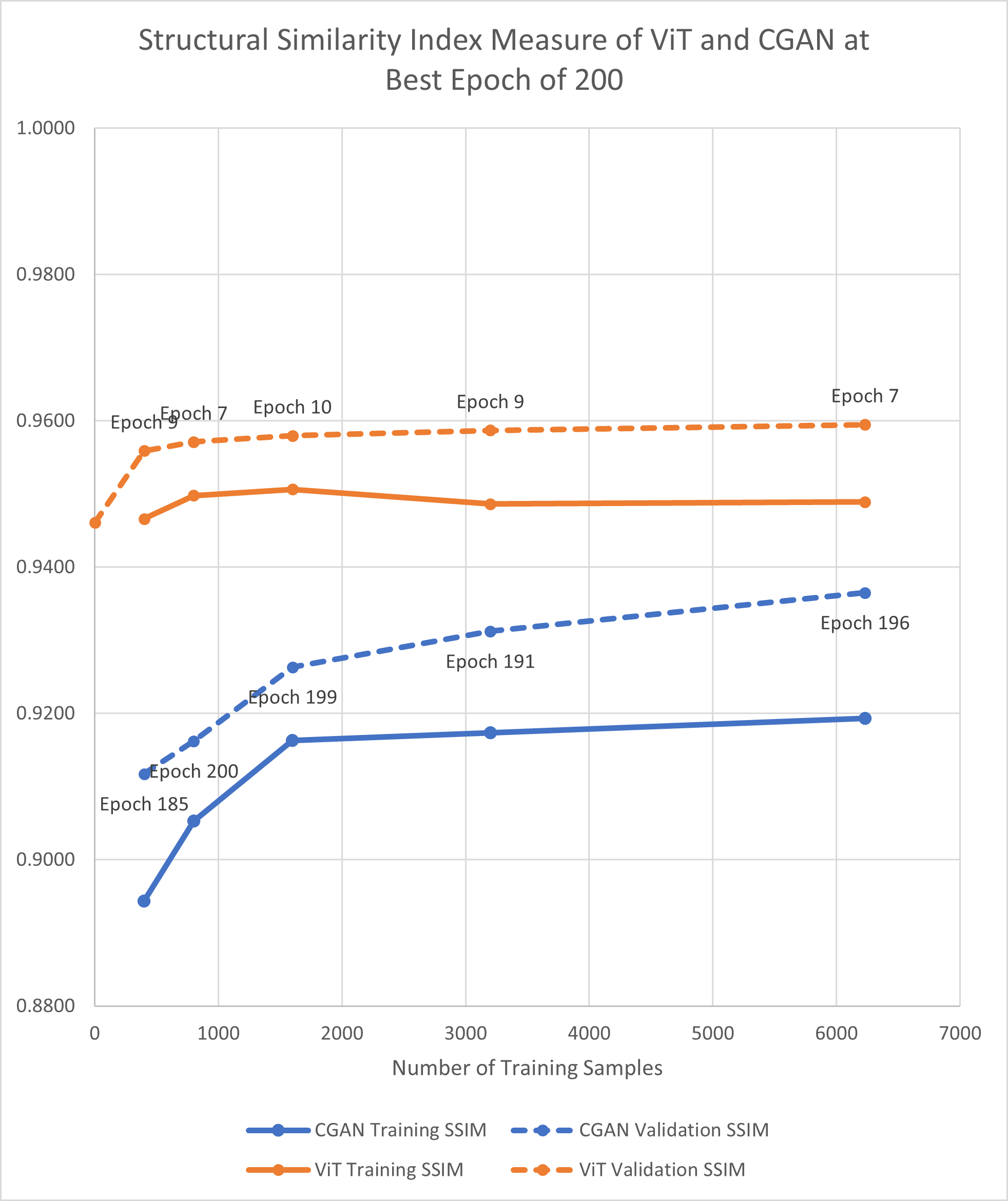}
}
\end{center}
\caption{MAE and SSIM for best epoch of 200 for E2 experiments (applying masks in all time steps). Best epoch is the best performance out of 5 runs for all experiments across all epochs.}
\label{round2-graphs}
\end{figure}

\begin{figure}[h]
\begin{center}
\fbox{
\includegraphics[width=12cm]{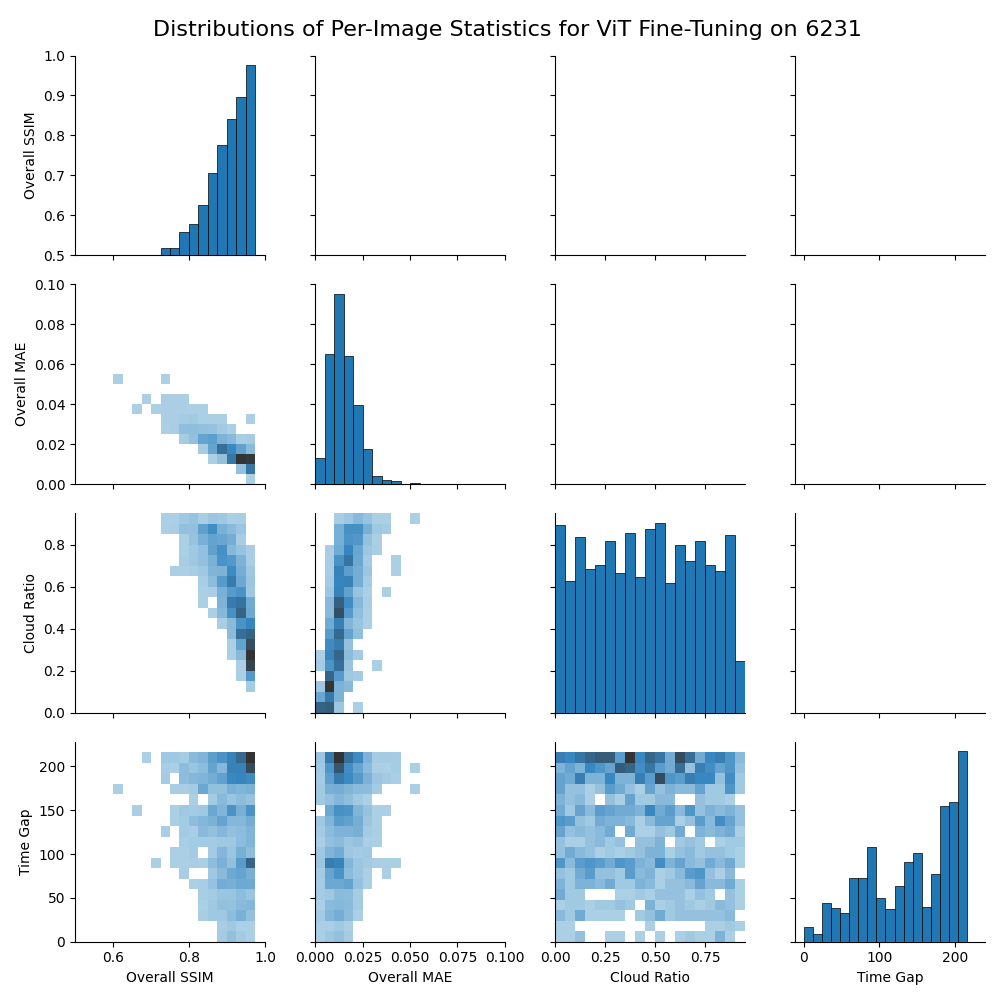}
}
\end{center}
\caption{Relationships between cloud cover, time gap, SSIM, and MAE of validation chips for best results of E1 experiments using the full dataset to fine-tune Prithvi. Statistics are calculated for each validation chip.}
\label{round1-vit-correlations}
\end{figure}

\begin{figure}[h]
\begin{center}
\fbox{
\includegraphics[width=12cm]{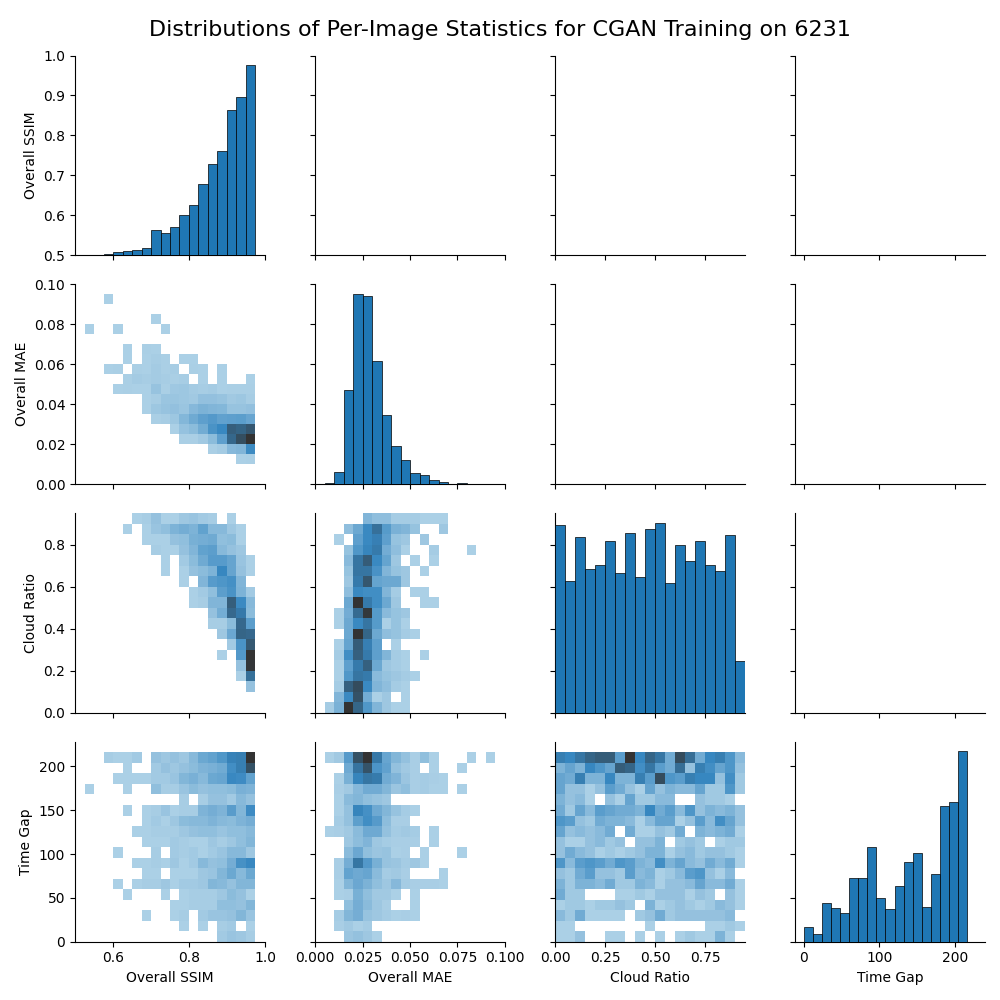}
}
\end{center}
\caption{Relationships between cloud cover, time gap, SSIM, and MAE of validation chips for best results of E1 experiments using the full dataset to train the CGAN. Statistics are calculated for each validation chip.}
\label{round1-cgan-correlations}
\end{figure}

\begin{figure}[h]
\begin{center}
\fbox{
\includegraphics[width=12cm]{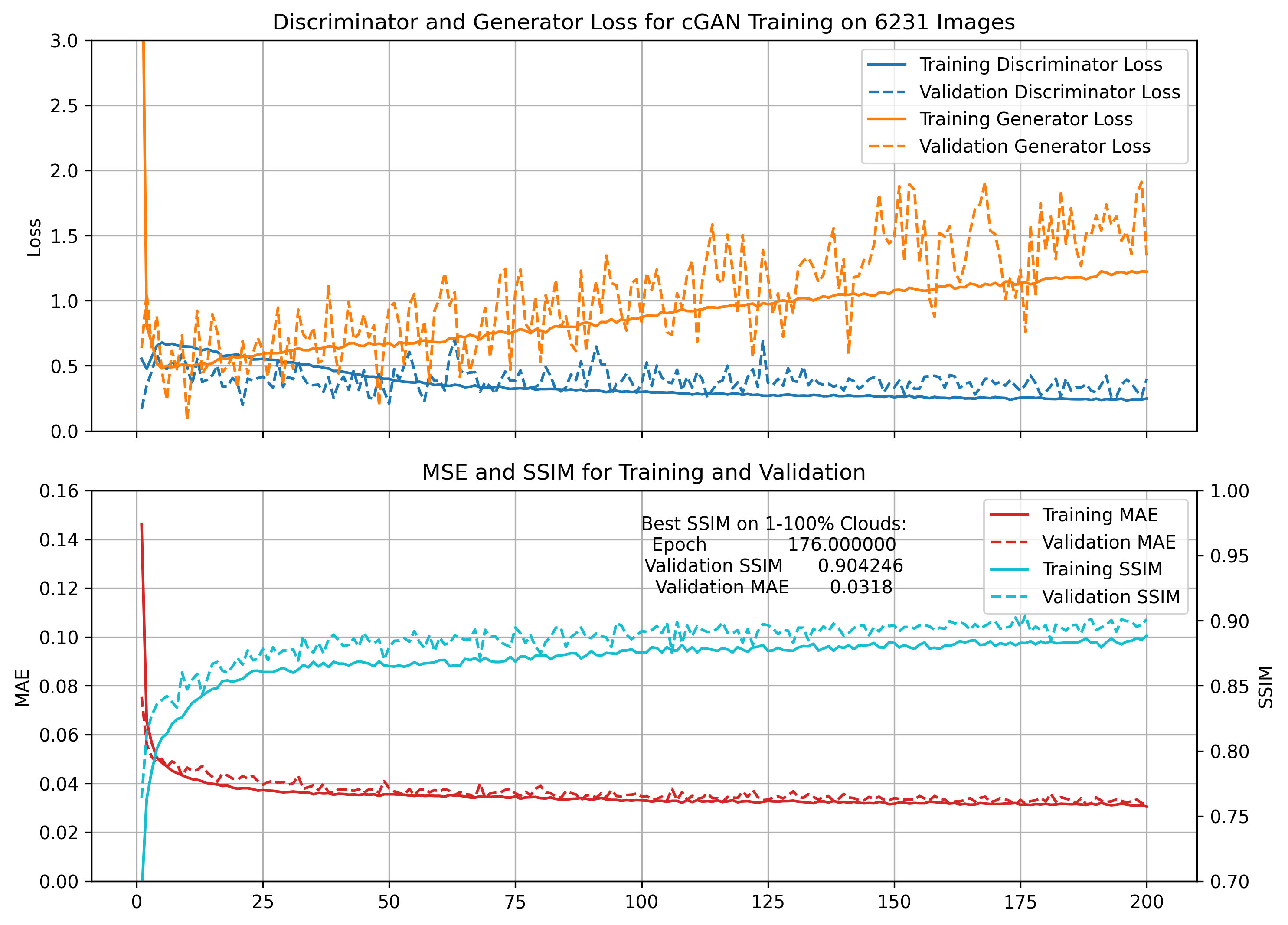}
}
\end{center}
\caption{Performance of the CGAN over the course of training on 6231 images in E1 experiments (applying mask to the middle scene)}
\label{cgan_training_graph_round1}
\end{figure}

\begin{figure}[h]
\begin{center}
\fbox{
\includegraphics[width=12cm]{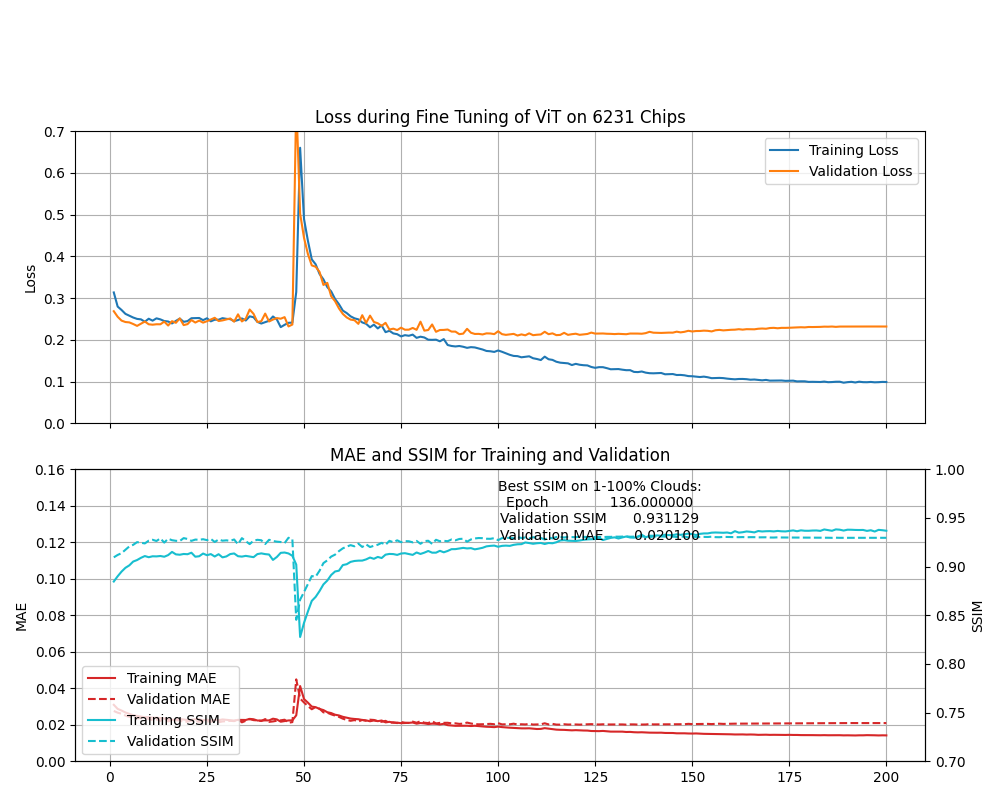}
}
\end{center}
\caption{Performance of Prithvi over the course of training on 6231 images in E1 experiments (applying mask to the middle scene)}
\label{vit_training_graph_round1}
\end{figure}

\begin{figure}[h]
\begin{center}
\fbox{
\includegraphics[width=12cm]{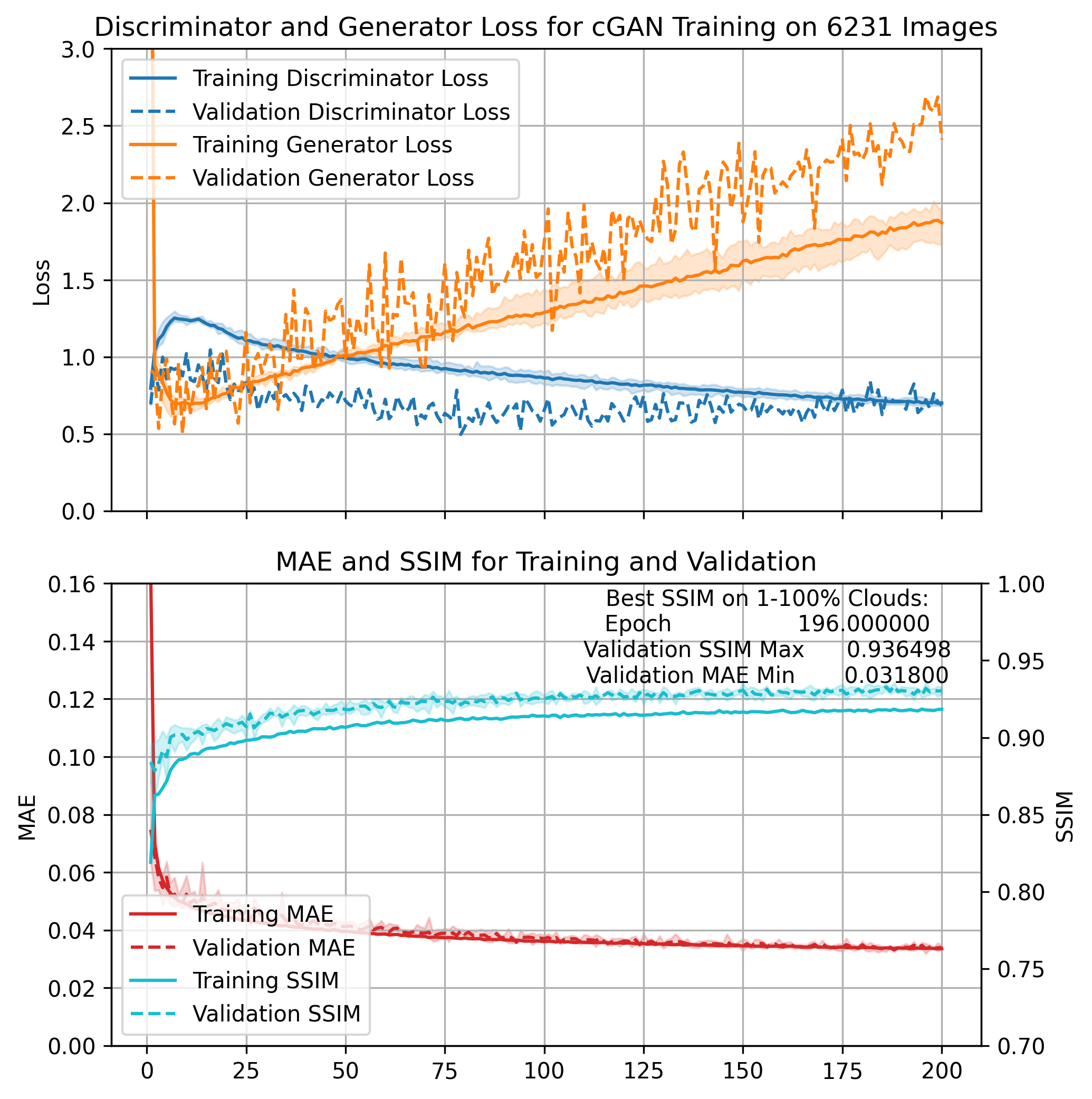}
}
\end{center}
\caption{Performance of the CGAN over the course of training on 6231 images in E2 experiments (applying masks in all time steps)}
\label{cgan_training_graph_round2}
\end{figure}

\begin{figure}[h]
\begin{center}
\fbox{
\includegraphics[width=12cm]{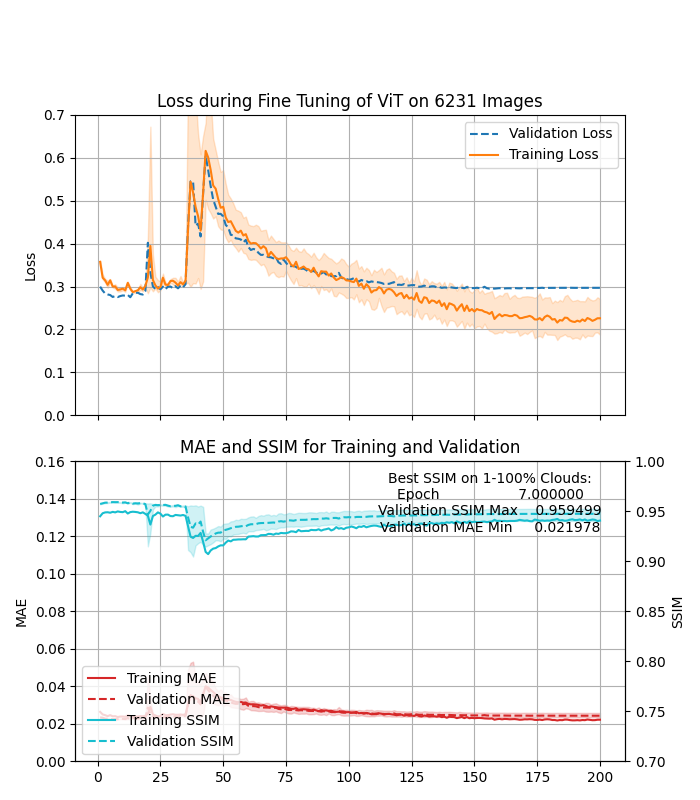}
}
\end{center}
\caption{Performance of Prithvi over the course of training on 6231 images in E2 experiments (applying masks in all time steps)}
\label{vit_training_graph_round2}
\end{figure}

\begin{figure}[h]
\begin{center}
\fbox{\includegraphics[width=10cm]{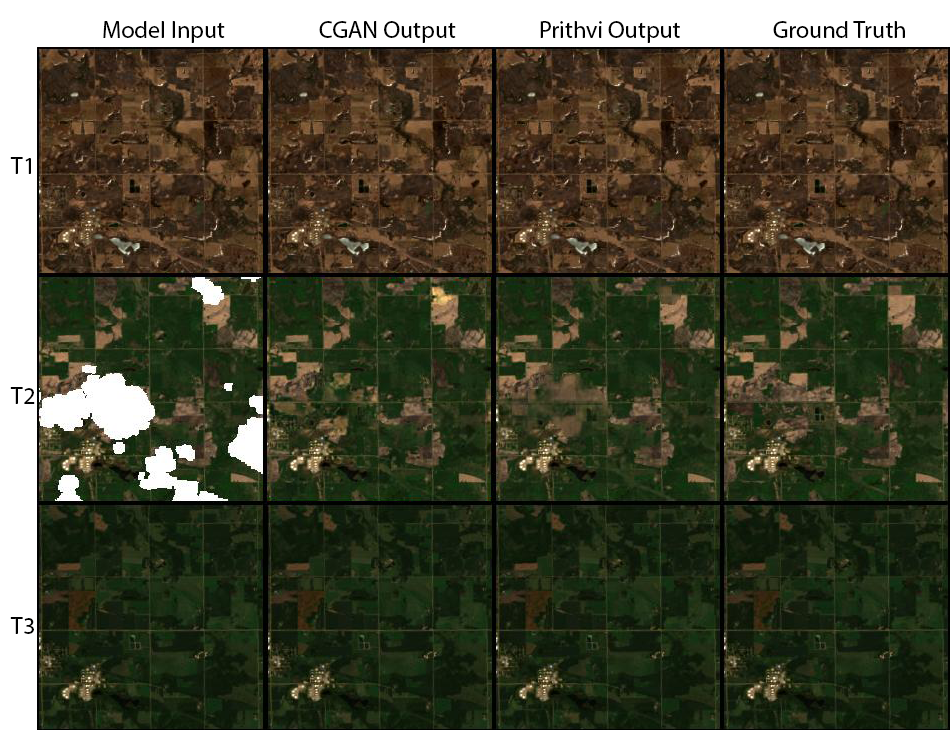}}
\end{center}
\caption{Reconstruction of a low-coverage image using CGAN and Prithvi, both trained using 6,231 images from E1 experiments (applying mask to the middle scene)}
\label{round1-lowcover}
\end{figure}

\begin{figure}[h]
\begin{center}
\fbox{\includegraphics[width=10cm]{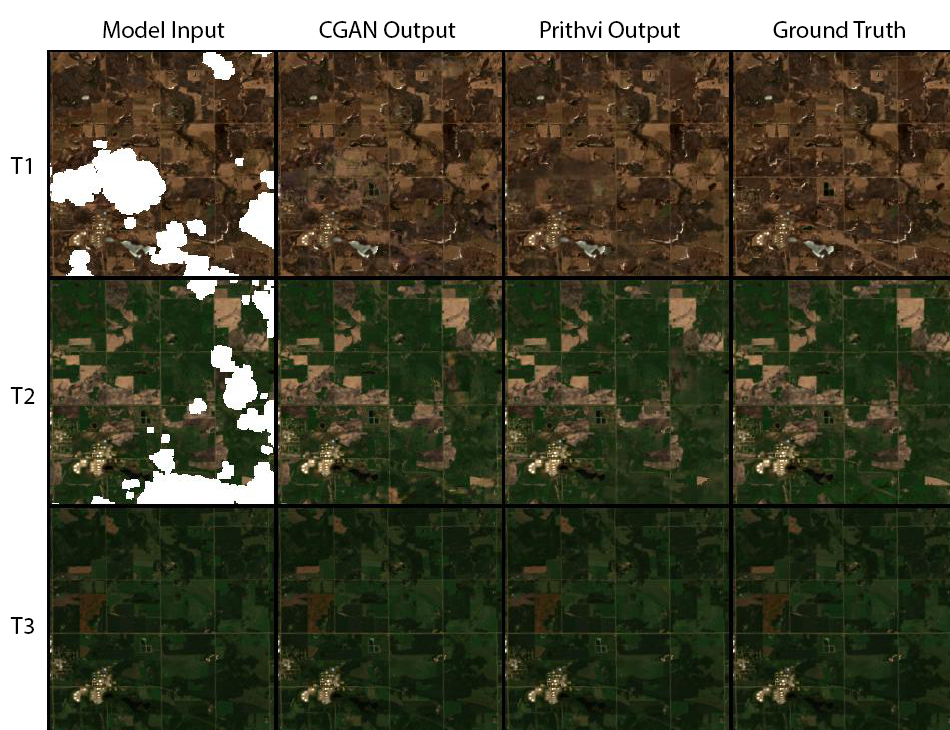}}
\end{center}
\caption{Reconstruction of a low-coverage image using CGAN and Prithvi, both trained using 6,231 images, from E2 experiments (applying masks in all time steps)}
\label{round2-lowcover}
\end{figure}

\begin{figure}[h]
\begin{center}
\fbox{\includegraphics[width=10cm]{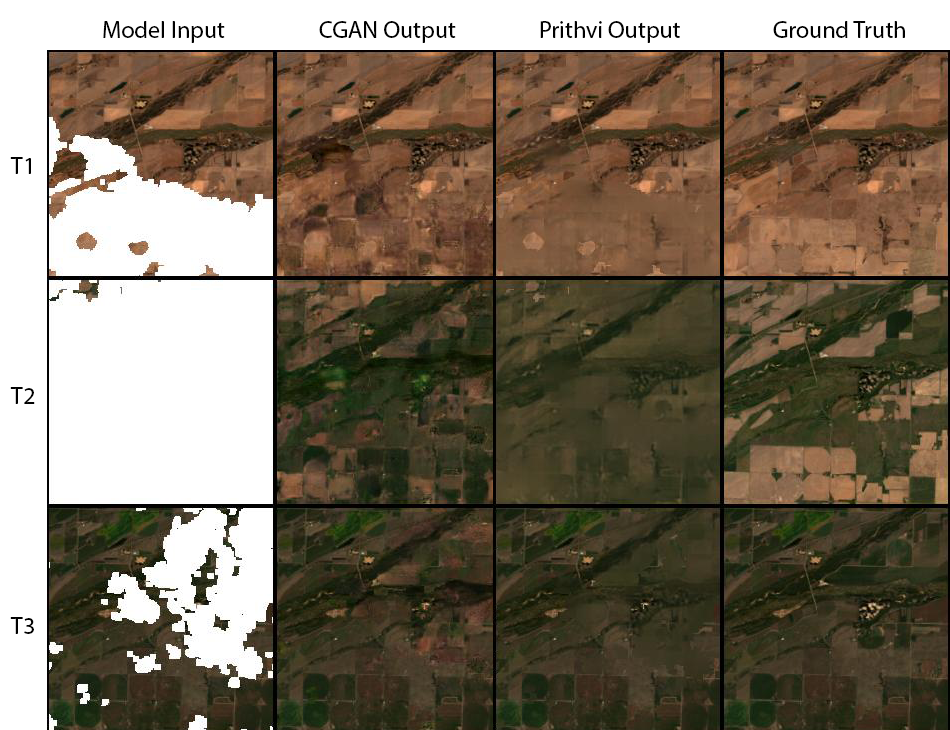}}
\end{center}
\caption{Reconstruction of high-coverage image using CGAN and Prithvi, both trained using 6,231 images, from E2 experiments (applying masks in all time steps)}
\label{round2-highcover}
\end{figure}

\begin{figure}[h]
\begin{center}
\fbox{\includegraphics[width=10cm]{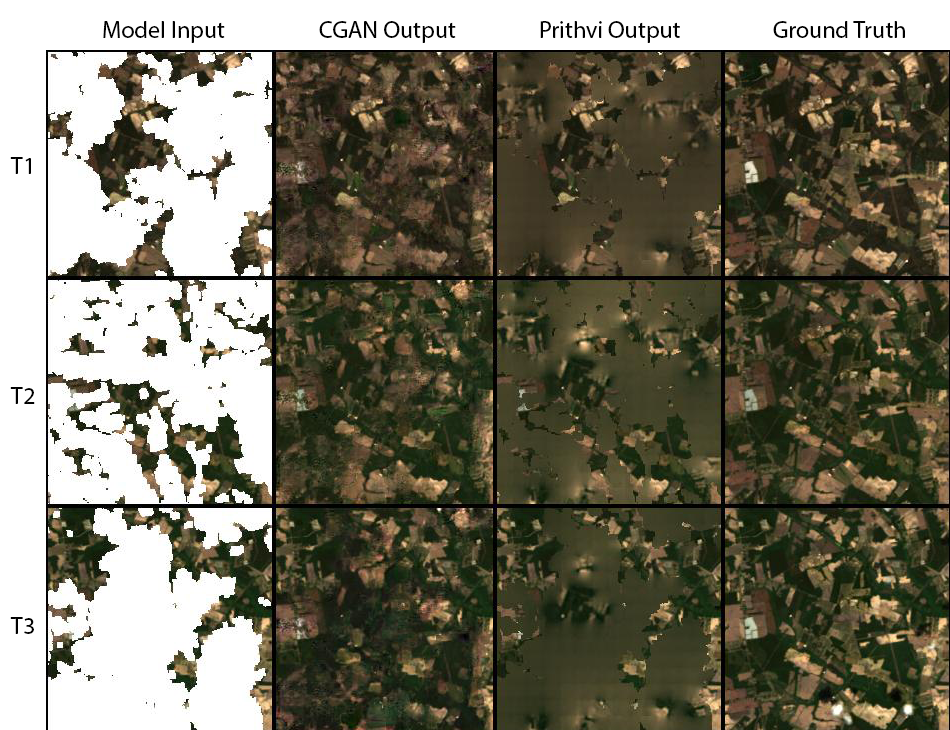}}
\end{center}
\caption{Reconstruction of a nearly full-coverage image using CGAN and Prithvi, both trained using 6,231 images from E2 experiments (applying masks in all time steps)}
\label{round2-fullcover}
\end{figure}

\begin{figure}[h]
\begin{center}
\fbox{\includegraphics[width=6cm]{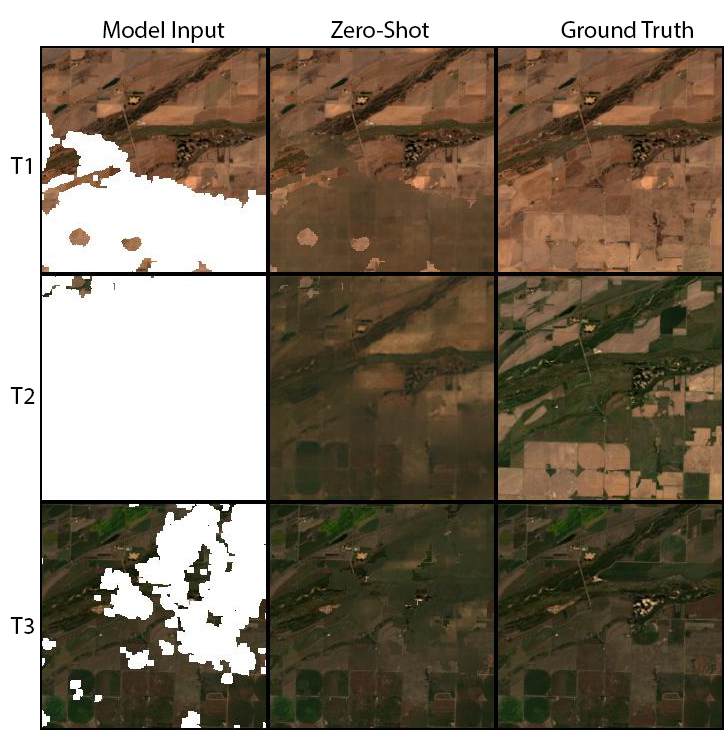}
\includegraphics[width=6cm]{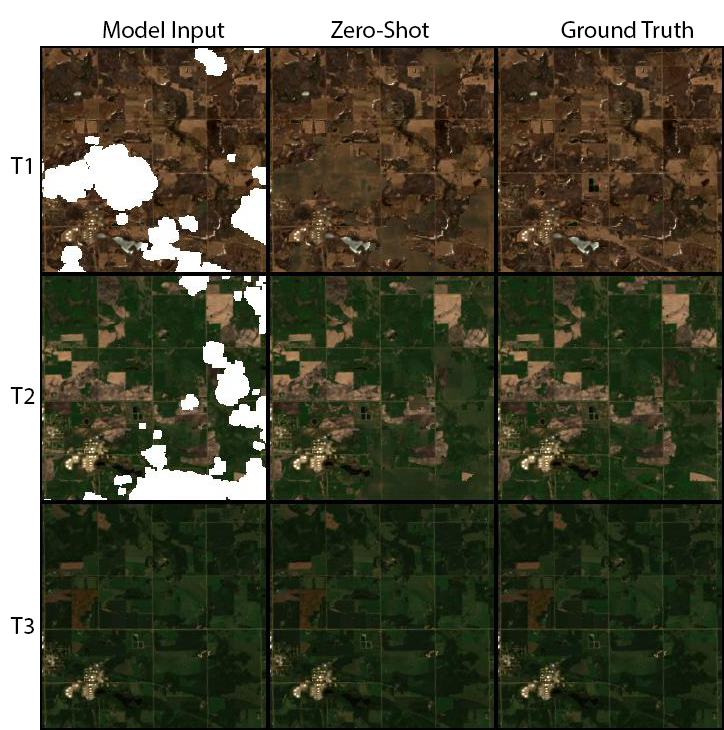}}
\end{center}
\caption{Reconstruction of a high and low coverage image using Prithvi with no fine-tuning from E2 experiments (applying masks in all time steps)}
\label{zeroshot_vis}
\end{figure}

\begin{figure}[h]
\begin{center}
\fbox{\includegraphics[width=10cm]{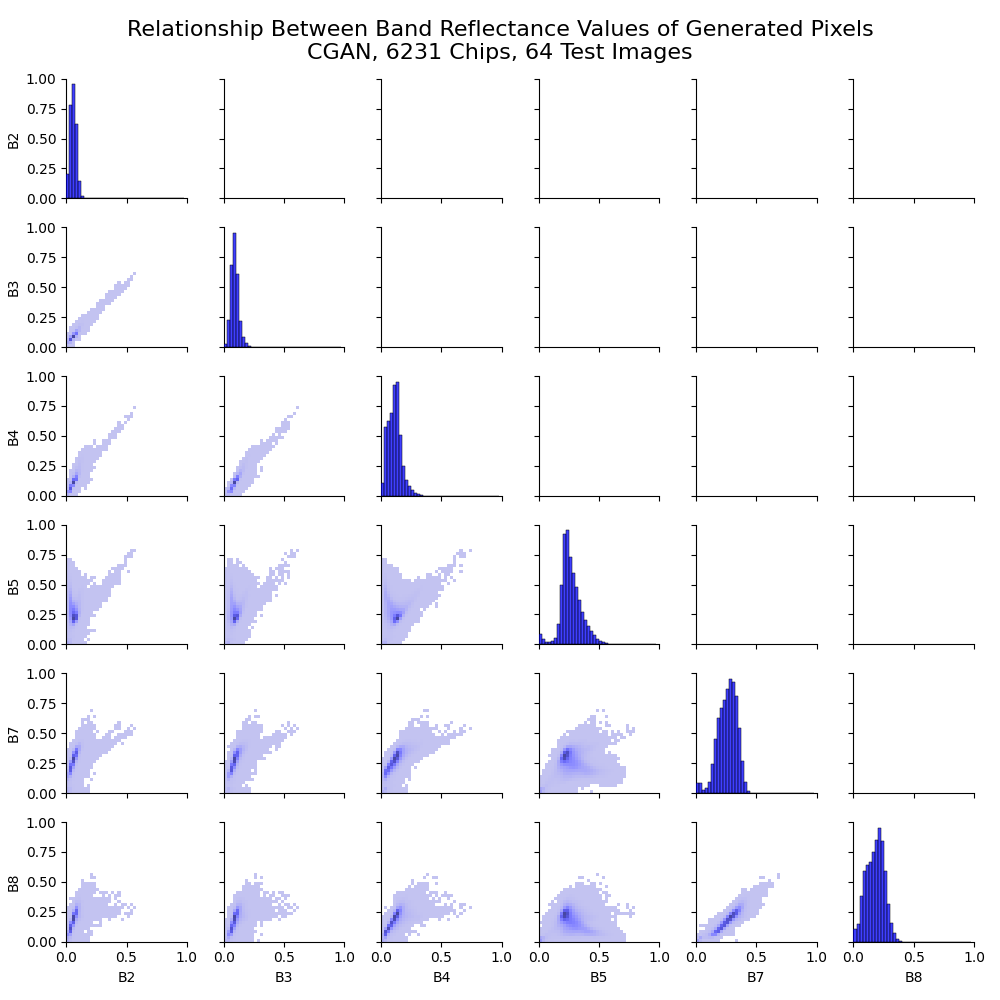}}
\fbox{\includegraphics[width=10cm]{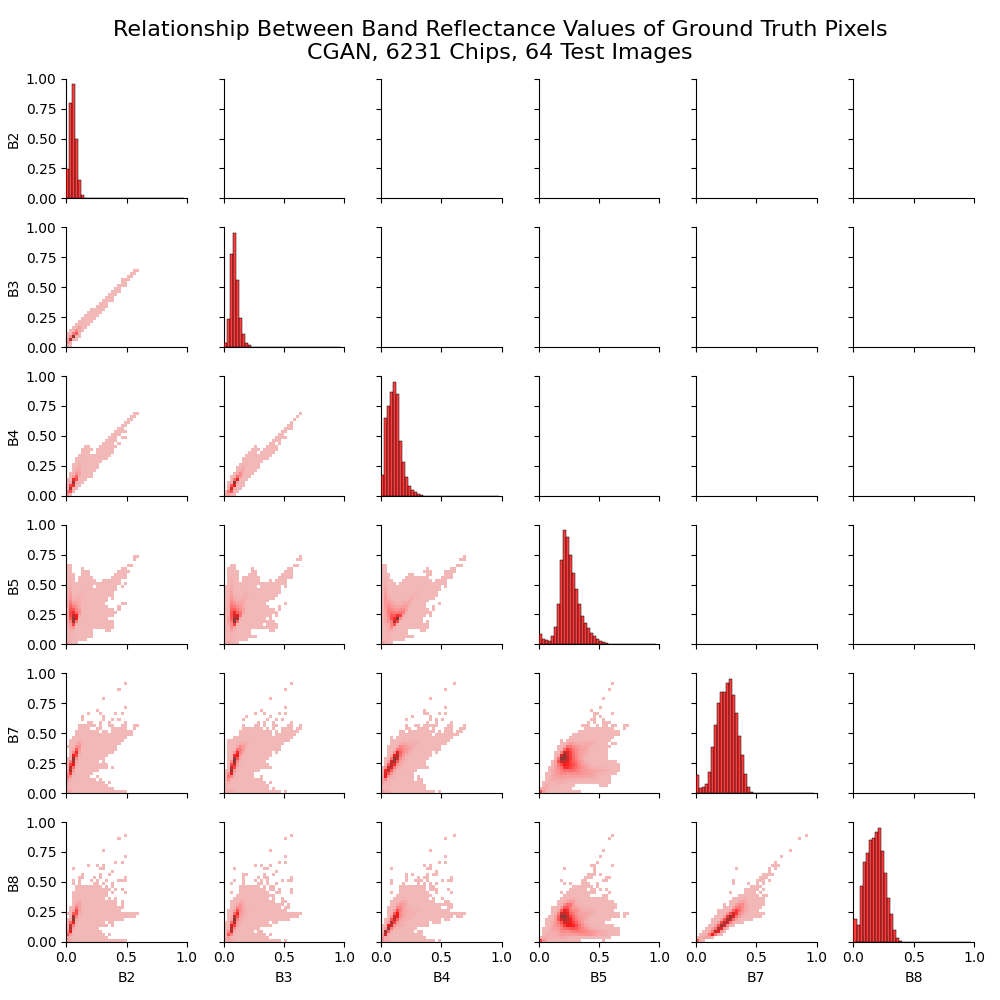}}
\end{center}
\caption{Band correlation matrices for the best CGAN model trained on the full dataset for E2 experiments (applying masks in all time steps). Band indices are shown for the first 64 image/mask combinations from the validation dataset, with generated values (top) and ground truth values (bottom).}
\label{round2-cgan-correlation}
\end{figure}

\begin{figure}[h]
\begin{center}
\fbox{\includegraphics[width=10cm]{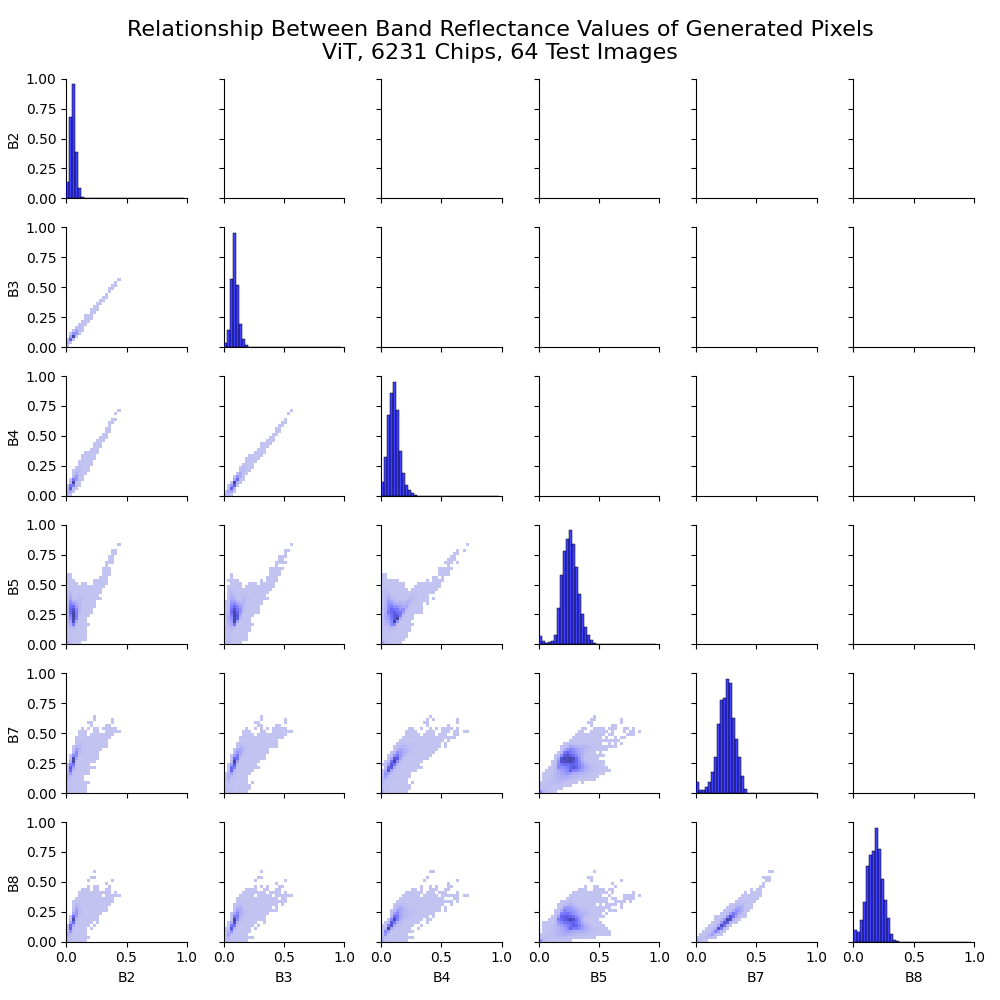}}
\fbox{\includegraphics[width=10cm]{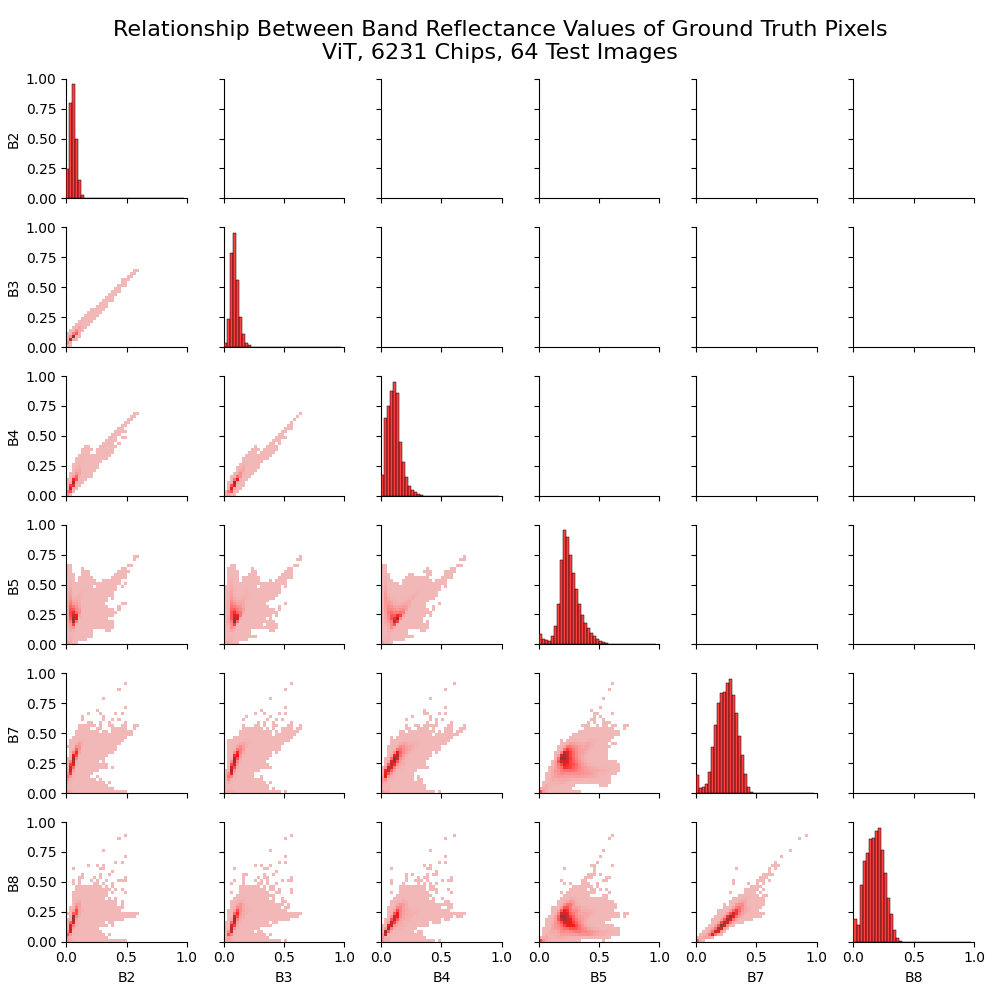}}
\end{center}
\caption{Band correlation matrices for the best fine-tuned Prithvi model trained on the full dataset for E2 experiments (applying masks in all time steps). Band indices are shown for the first 64 image/mask combinations from the validation dataset, with generated values (top) and ground truth values (bottom).}
\label{round2-vit-correlation}
\end{figure}

\begin{figure}[h]
\begin{center}
\fbox{\includegraphics[width=10cm]{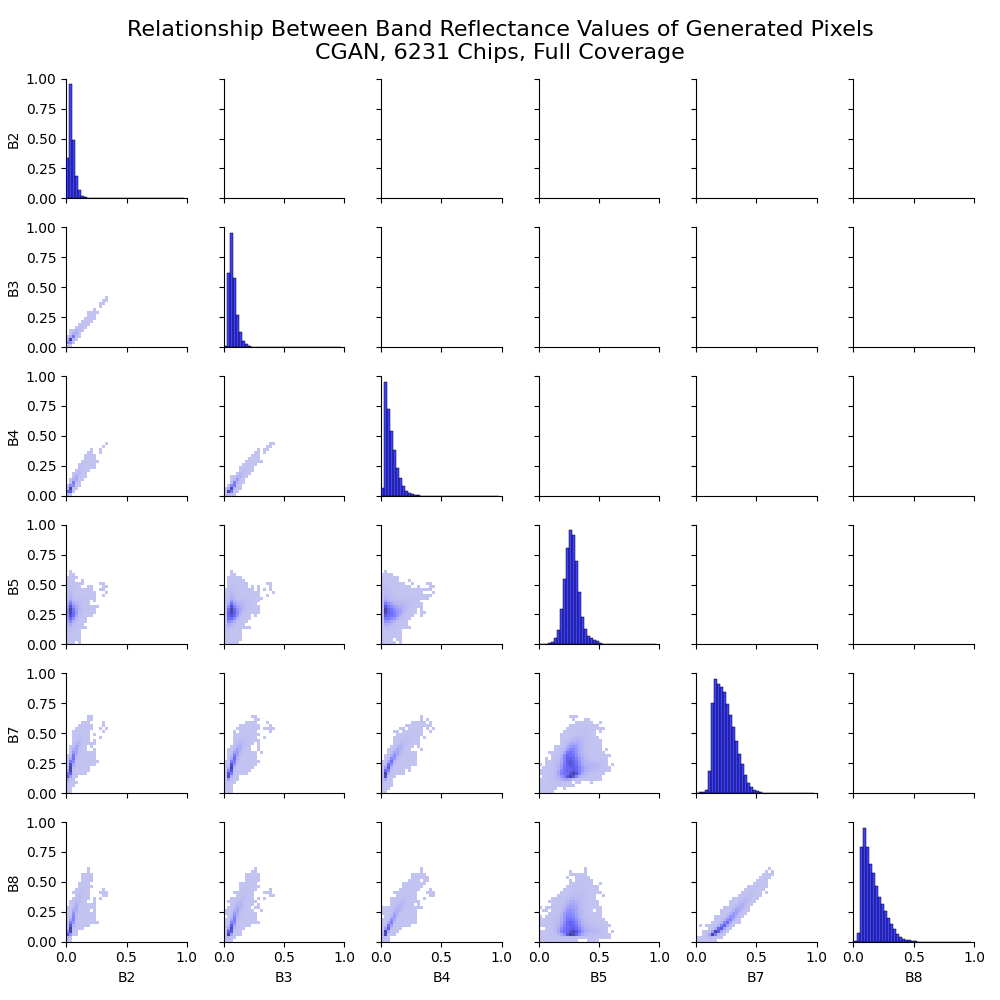}}
\fbox{\includegraphics[width=10cm]{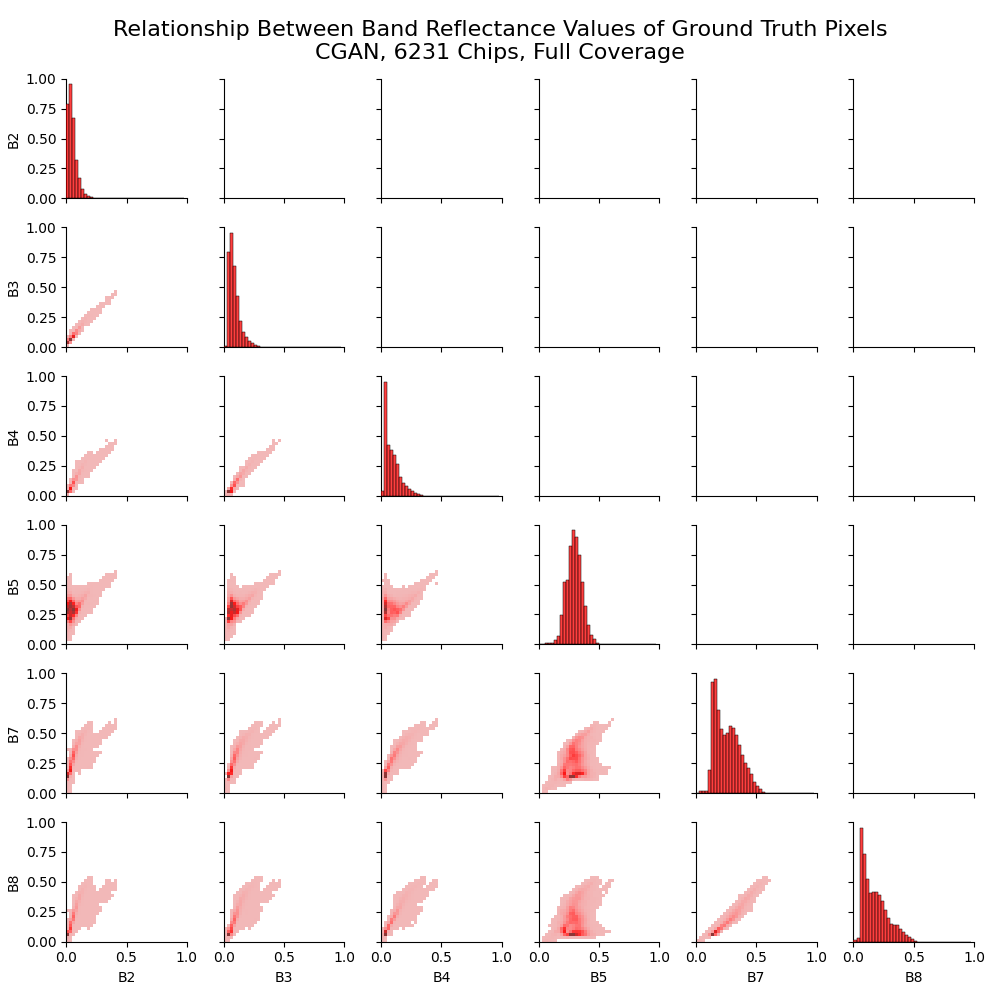}}
\end{center}
\caption{Band correlation matrices for the best fine-tuned CGAN model trained on the full dataset for E2 experiments (applying masks in all time steps). Band indices are shown for the sample image in \ref{round2-fullcover}, with generated values (top) and ground truth values (bottom).}
\label{round2-cgan-correlation-full}
\end{figure}

\begin{figure}[h]
\begin{center}
\fbox{\includegraphics[width=10cm]{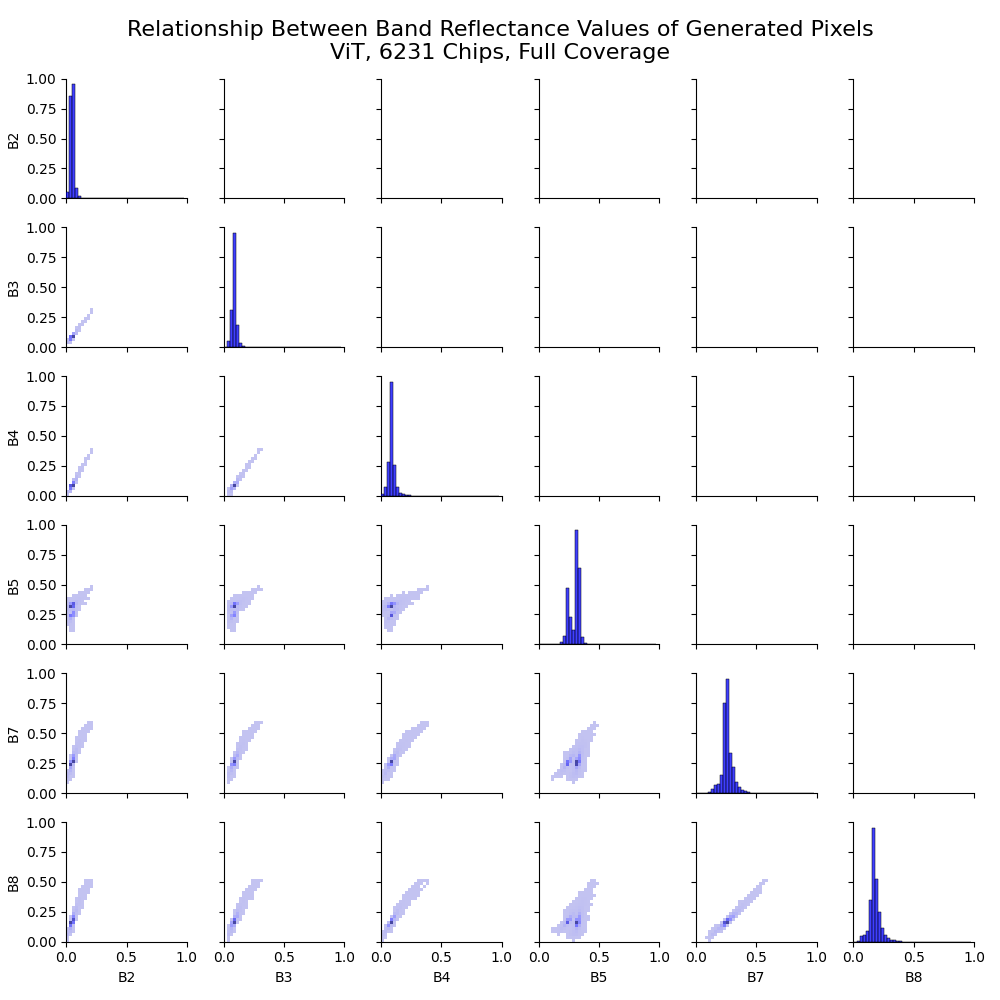}}
\fbox{\includegraphics[width=10cm]{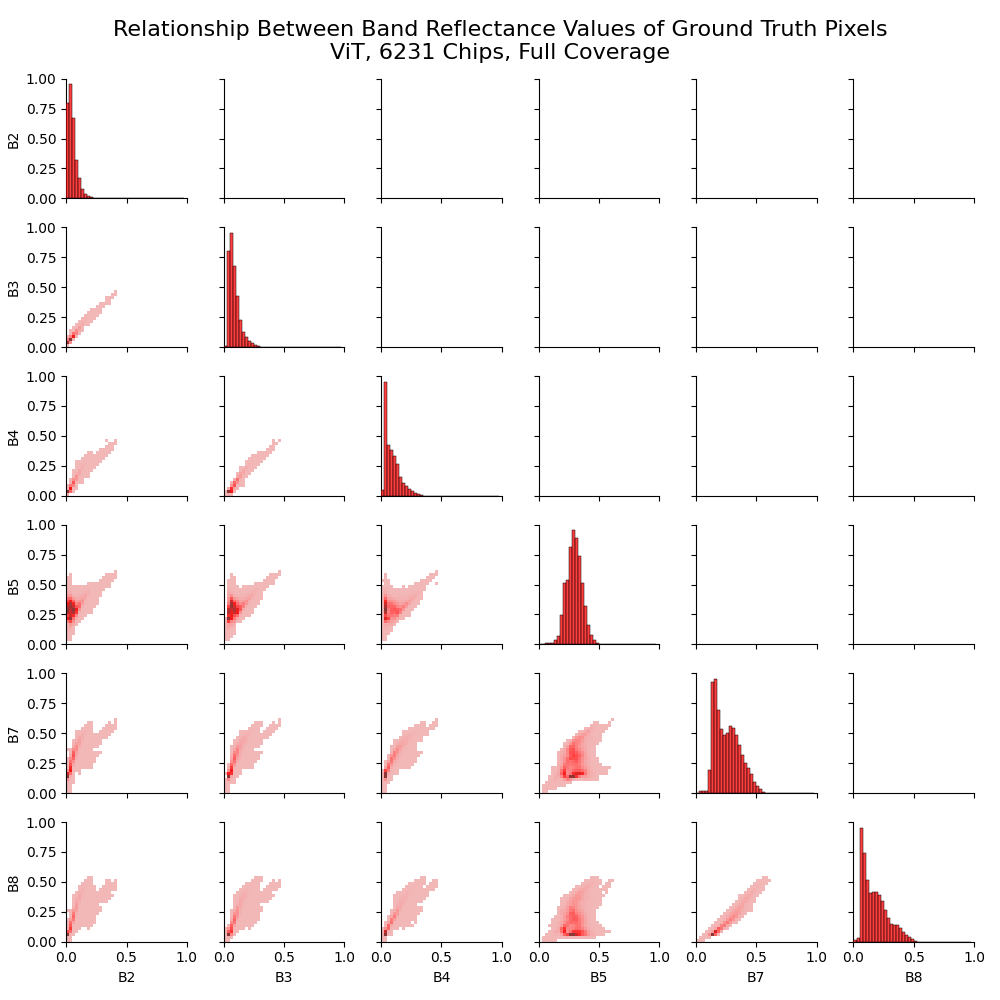}}
\end{center}
\caption{Band correlation matrices for the best fine-tuned Prithvi model trained on the full dataset for E2 experiments (applying masks in all time steps). Band indices are shown for the sample image in \ref{round2-fullcover}, with generated values (top) and ground truth values (bottom).}
\label{round2-vit-correlation-full}
\end{figure}

\end{document}